\documentclass[11pt]{article}

\usepackage[preprint]{acl}

\usepackage{times}
\usepackage{latexsym}

\usepackage[T1]{fontenc}

\usepackage[utf8]{inputenc}

\usepackage{microtype}

\usepackage{url}
\usepackage{listings}
\usepackage{tabularx}
\usepackage{multirow}
\usepackage{amsmath}
\usepackage{amsfonts}
\usepackage{multirow}
\usepackage{graphicx} 
\usepackage{booktabs}

\newcommand{\gain}[1]{{\scriptsize\color{gray}(#1)}}

\usepackage{inconsolata}

\usepackage{graphicx}

\usepackage[colorinlistoftodos,textsize=tiny, disable]{todonotes}

\title{CaseFacts: A Benchmark for Legal Fact-Checking and Precedent Retrieval}

\author{Akshith Reddy Putta, Jacob Devasier, Chengkai Li \\
  University of Texas at Arlington \\
  \texttt{\{akshith.putta, jacob.devasier, cli\}@uta.edu} \\}

\begin{document}
\maketitle
\begin{abstract}
Automated Fact-Checking has largely focused on verifying general knowledge against static corpora, overlooking high-stakes domains like law where truth is evolving and technically complex. We introduce \textbf{CaseFacts}, a benchmark for verifying colloquial legal claims against U.S. Supreme Court precedents. Unlike existing resources that map formal texts to formal texts, CaseFacts challenges systems to bridge the semantic gap between layperson assertions and technical jurisprudence while accounting for temporal validity. The dataset consists of 6,294 claims categorized as \textsc{Supported}, \textsc{Refuted}, or \textsc{Overruled}. We construct this benchmark using a multi-stage pipeline that leverages Large Language Models (LLMs) to synthesize claims from expert case summaries, employing a novel semantic similarity heuristic to efficiently identify and verify complex legal overrulings. Experiments with state-of-the-art LLMs reveal that the task remains challenging; notably, augmenting models with unrestricted web search degrades performance compared to closed-book baselines due to the retrieval of noisy, non-authoritative precedents. We release CaseFacts\footnote{\url{https://github.com/idirlab/CaseFacts}} to spur research into legal fact verification systems.
\end{abstract}

\section{Introduction}

Automated Fact-Checking (AFC) has emerged as a critical safeguard against misinformation, yet existing systems primarily operate within general-knowledge domains, validating claims against corpora like Wikipedia~\cite{Aly2021FEVEROUSFE, akhtar-etal-2022-pubhealthtab}. However, high-stakes domains such as law present unique challenges that render general-purpose AFC systems inadequate. In the legal sphere, verification requires more than surface-level textual overlap; it demands the retrieval of authoritative precedent, the interpretation of complex holdings, and the temporal awareness to determine if a ruling is still valid.

The primary bottleneck in Legal AFC is the semantic gap between public discourse and judicial writing. While most people assert claims in colloquial language (e.g., ``The Supreme Court banned school prayer''), the supporting evidence exists in formal, technical jurisprudence (e.g., ``state officials may not compose an official state prayer''). Current legal benchmarks, such as LegalBench~\cite{guha2023legalbenchcollaborativelybuiltbenchmark} and CaseHOLD~\cite{zheng2021doespretraininghelpassessing}, facilitate legal reasoning and judgment prediction but operate strictly within the formal register---mapping legalese to legalese. They do not evaluate a model's ability to bridge the linguistic disparity between an informal claim and a formal ruling~\cite{chen2025-retrieving, mori2025assessingperformancegaplexical}.

Furthermore, legal truth is temporally fragile. A claim that was true in 1990 may be false today if the underlying precedent has been overruled. Existing datasets rarely explicitly model this validity constraint, often treating legal documents as static text rather than a dynamic system of evolving rules.

To address these limitations, we introduce \textbf{CaseFacts}, a new benchmark dataset designed to evaluate the verification of colloquial claims against U.S. Supreme Court (SCOTUS) rulings. Unlike previous resources, CaseFacts is constructed to test three specific dimensions of legal verification: (1) {Retrieval across the semantic gap}, mapping informal assertions to technical case law; (2) {Reasoning against hard negatives}, distinguishing true principles from plausible but factually refuted distortions; and (3) {Precedent changes over time}, identifying when a claim relies on a precedent that has been explicitly \textsc{Overruled}.

Constructing such a dataset presents significant challenges due to the scarcity of expert-annotated data. We introduce a rigorous, multi-stage pipeline leveraging Large Language Models (LLMs) to synthesize claims from expert summaries of 3,299 SCOTUS cases sourced from Oyez\footnote{\url{https://www.oyez.org/}}. Our pipeline generates diverse claim-evidence pairs and employs a novel LLM-as-a-Judge verification loop to filter for factuality, resolve internal contradictions, and ensure stylistic diversity. To validate the \textsc{Overruled} class, we implement a heuristic filtering method based on semantic similarity intervals to efficiently locate inter-case contradictions, which are then verified against case metadata. The final dataset consists of 6,294 claims, including a high-quality, human-annotated test set.

We benchmark state-of-the-art LLMs and retrieval models on CaseFacts, yielding several critical insights. First, we find that allowing LLMs unrestricted web search does not necessarily improve performance; in our baselines, a search-enabled GPT-4o performed worse than a naive, closed-book setting, often retrieving noisy or tangentially relevant cases. Second, we demonstrate that while standard Natural Language Inference (NLI) models fail to grasp the nuances of legal overrulings, embedding models fine-tuned on our training set significantly improve retrieval recall. 

In summary, our contributions are as follows:
\begin{itemize}
    \item We introduce \textbf{CaseFacts}, the first large-scale benchmark for verifying colloquial legal claims against SCOTUS precedents, featuring distinct \textsc{Supported}, \textsc{Refuted}, and \textsc{Overruled} classes.
    \item We propose a robust synthetic data generation pipeline that resolves intra-case and inter-case inconsistencies, using semantic similarity heuristics to efficiently identify legal overrulings and confirmations in large corpora.
    \item We provide comprehensive benchmarks demonstrating that CaseFacts poses a significant challenge to current LLMs, particularly in bridging the semantic gap and retrieving the precise controlling authority required for legal verification.
\end{itemize}

\section{CaseFacts Benchmark}

\subsection{Task Formulation} 
We formulate legal fact-checking as a combined retrieval and verification task. Given a colloquial legal claim $c$ and a knowledge base of case law $\mathcal{K}$, the system must output a verdict $y \in \{\textsc{Supported}, \textsc{Refuted}, \textsc{Overruled}\}$ and a set of supporting evidence SCOTUS cases $E \subset \mathcal{K}$. A claim is \textsc{Supported} if it is entailed by the holding of a valid case in $\mathcal{K}$. It is \textsc{Refuted} if it contradicts the holding of a valid case. It is \textsc{Overruled} if it relies on a holding from a case $k_{old} \in \mathcal{K}$ that has been explicitly overturned by a subsequent reversing case $k_{new} \in \mathcal{K}$.

\subsection{Data Source}
Our primary source for legal documentation is Oyez, a multimedia archive of the US Supreme Court. We utilized the ``facts,'' ``question,'' and ``conclusion'' fields for each case, each of which are summaries synthesized by legal experts using original Supreme Court documentation. We refer to these fields as case ``evidence''. These expert summaries provide a high-quality, structured foundation for generating synthetic legal claims. We filtered the corpus to include only cases containing all three fields, resulting in a collection of 3,299 cases (as of July 2025).

\subsection{Dataset Creation Pipeline}
\begin{figure*}[t]
  \centering
  \includegraphics[width=0.93\linewidth]{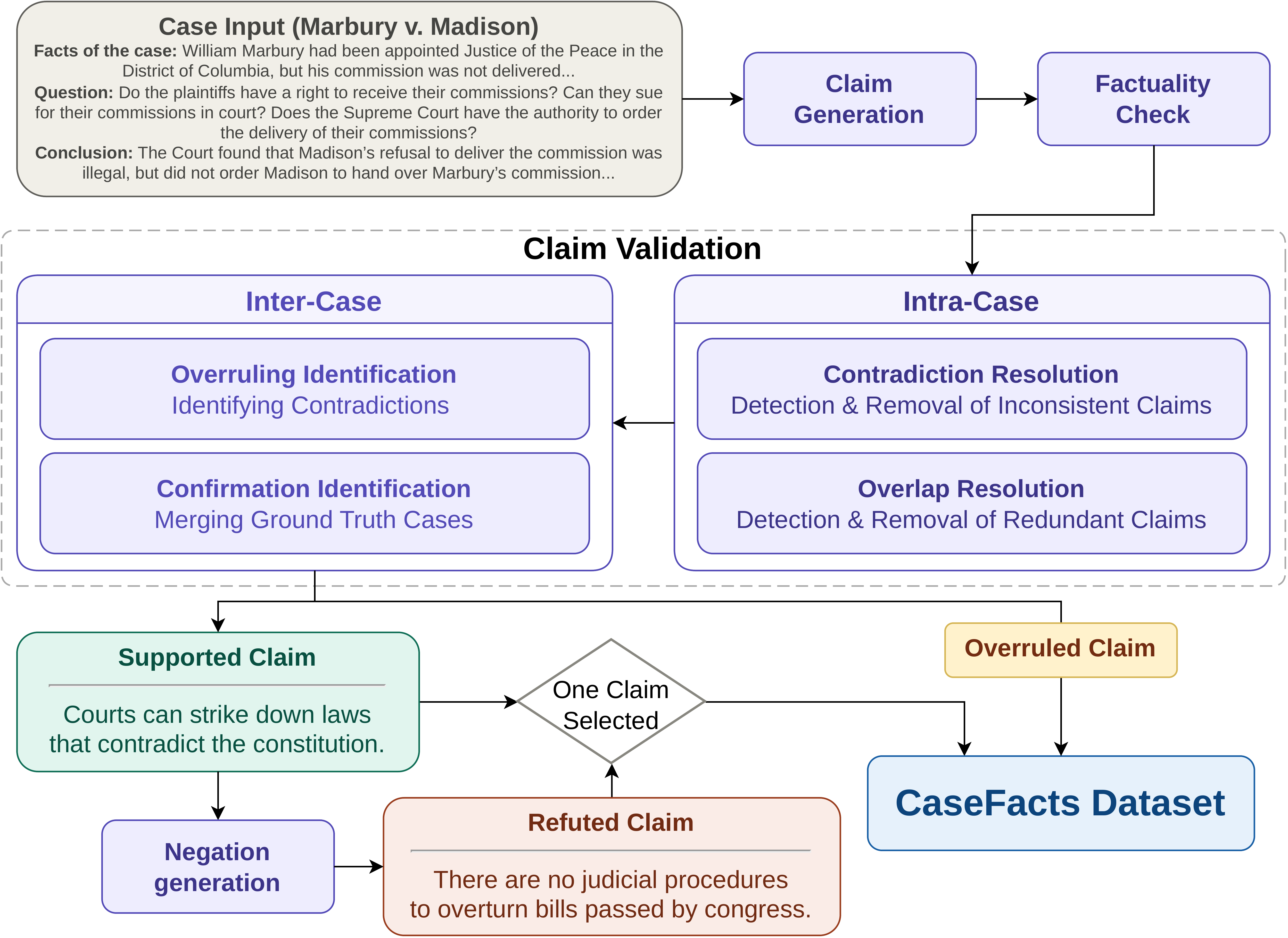}
  \caption{\label{fig:pipeline-example} Dataset creation pipeline with the example of Marbury v. Madison and claims of each class. This case, famous for establishing judicial review, was ruling over the delivery of commissions, not judicial review itself. The pipeline, taking as a case as input, generates and validates claims based on the case. The pipeline's steps using an LLM are in purple. The Claim Generation step uses a different LLM than the other steps.}
\end{figure*}

We implemented a multi-stage pipeline (Figure \ref{fig:pipeline-example}) to generate synthetic claims and ensure their factuality, consistency, and diversity. To mitigate self-preference bias during validation~\cite{wataoka2025selfpreferencebiasllmasajudge}, we utilized different model families for generation (Qwen3-Next-80B-A3B-Thinking)~\cite{qwen3technicalreport} and post-generation validation (gemma-3-27b-it)~\cite{gemmateam2025gemma3technicalreport}. We use open-weight models only to promote reproducibility of our dataset creation method. For all LLMs, temperature was set to 0.6 and top-p to 0.9.

\subsubsection{Claim Generation}
\label{sec:claim-gen}
We prompted the generation model with the case evidence from each source case and instructed it to generate claims derived from the case's legal principles (Prompt~\ref{lst:claim_gen_prompt}). To ensure the retrieval task remained non-trivial and generalizable, we explicitly constrained the model to exclude case-specific identifiers, such as party names or dates. Furthermore, we enforced strict stylistic constraints prohibiting ``court-report phrasing'' (e.g., ``The court found that...''). This ensures the output consists of direct, atomic legal assertions rather than summaries of judicial opinions. This process yielded an initial pool of 10,471 unique claims.

\subsubsection{Intra-Case Validation}
We applied filters to ensure factual consistency and minimize redundancy among claims derived from the same case. These inconsistencies would hinder the creation of a high-quality dataset.

\paragraph{Factuality Check} We employed an LLM-as-a-judge to verify that each claim was strictly entailed by the source case evidence (Prompt~\ref{lst:factuality_prompt}). We also explicitly instructed the model to adhere to a closed-world assumption, prohibiting reliance on outside legal knowledge or assumptions not present in the provided case summary. This process removed 1,484 claims that failed to meet the expected entailment criteria, resulting in a pool of 8,987 factually consistent claims.

\paragraph{Claim Inconsistencies} An issue that was noticed was the presence of the LLM creating factually-inconsistent claims for claims within the same case. These are errors associated with LLM generation. We refer to these as ``contradictions,'' even though most of them are smaller factual inconsistencies with each other when relating to the evidence of the case. Another issue that was noticed was the presence of very similar, redundant claims generated from the same case, and we refer to these as ``overlaps.'' This is possibly due to the LLM generating unnecessary claims, although it has been instructed not to. The aim of the dataset is to generate claims which have a ``legal principle'' from the case that is precedent. Therefore, redundant claims must be removed to maintain a dataset consisting of high-quality, meaningfully-different claims. We collectively refer to overlaps and contradictions as claim inconsistencies.

\paragraph{Intra-Case Inconsistency Resolution} We detected 90 pairs of internally contradictory claims and 1,568 pairs of semantically redundant claims using Prompts~\ref{lst:contra_same_case_prompt} and~\ref{lst:overlap_same_case_prompt}, for pairs within the same case. The smaller number of contradictory claims is possibly due to the factuality check removing most of the inconsistent claims, with the remaining identified with this pass. A resolution step (Prompts~\ref{lst:contra_same_case_resolution_prompt}, \ref{lst:overlap_same_case_resolution_prompt}) removed one claim from each of their respective pairs. This removed 1,404 claims, resulting in a filtered pool of 7,583 claims.

\subsubsection{Inter-Case Validation}
We addressed inconsistencies across different cases to identify cross-case confirmations (claims supported by the rulings of multiple cases) and construct the \textsc{Overruled} class with the detected contradictions. A naive approach would involve checking all $N^2$ possible cross-case pairs for inconsistencies, but this requires over 25 million comparisons, which is computationally infeasible for LLMs, even smaller ones like gemma-3-27b-it.

\paragraph{Inadequacy of Standard NLI}
To filter these pairs efficiently, we first attempted to use a standard Natural Language Inference (NLI) model (\texttt{textattack/bert-base-uncased-MNLI}) to identify contradictions and entailments across the full dataset. However, qualitative analysis revealed this standard NLI model consistently performed very poorly in grasping the deeper legal entailment required to identify overrulings and confirmations, rendering its predictions unusable.

\paragraph{Similarity-Based Filtering}
Consequently, we adopted a heuristic approach based on semantic similarity (detailed in Section~\ref{sec:sim_dist_diff_case}). To determine the filtering thresholds, we drew a uniform sample of 1,000 pairs from each 0.1-width similarity interval of all claim pair combinations. Then, we applied LLM-as-a-judge to detect contradictions and overlaps in each bin. We then computed percentile statistics over the detected inconsistencies and selected the 95th‑percentile range as our filtering window, i.e., the interval capturing 95\% of all detected contradictions and overlaps. This yielded specific semantic similarity intervals of 0.70--0.95 for contradictions and 0.80--0.97 for overlaps (as detailed in Table~\ref{tab:sim-experiment}).

The intervals derived from the sampling distribution contained 199,203 pairs for contradictions and 15,874 pairs for overlaps. We therefore restricted the expensive LLM-based verification to pairs falling within these high-risk intervals, reducing the workload to a manageable subset while maintaining high recall.


\paragraph{Identifying Overrulings} We identified 1,946 contradicting pairs, 0.98\% of the sampled range. In the legal domain, such contradictions often signify that one case overrules another. We resolved these using Prompt~\ref{lst:contra_diff_case_res_prompt}, which provided the model with the ruling dates of both cases to determine temporal precedence. To minimize false positives, the prompt instructed the model to favor an interpretation of consistency where possible, reserving the \textsc{Overruled} label for clear legal overrulings. We then validated potential overrulings against Oyez disposition metadata. If the overruling case had a valid disposition (e.g., ``reversed'' or ``remanded''), we labeled the claim from the older case as \textsc{Overruled} and appended the newer case to its evidence. This process established the \textsc{Overruled} class, a distinct contribution of our benchmark.

\paragraph{Identifying Confirmations} For the overlap detection task, we used prompt~\ref{lst:overlap_diff_case_prompt}, yielding 12.31\% of the pairs, 1,955 samples. Prompt~\ref{lst:overlap_diff_case_res_prompt} was used for resolving the overlapping claim pairs. For the overlapping claim pairs, we have updated the ground truth cases of the remaining claim to consist of both cases from the claim pair that overlapped, as the claim is confirmed by multiple cases. If there is a conflict when merging ground truth cases, where one claim is to merge over or under other claims, we select the pair with the highest similarity and instruct the LLM to indicate if the other claim can be merged under this pair. If so, the 3rd claim's case will be included as part of the original claim pair's ground truth cases.

\begin{table}
\centering
\begin{tabularx}{\linewidth}{Xcc}
\toprule
\textbf{Category} & \textbf{Percentile} & \textbf{Similarity Range} \\
\midrule
Contradictions & 50\%  & 0.820 -- 0.918 \\
Contradictions & 68\%  & 0.809 -- 0.922 \\
Contradictions & 75\%  & 0.805 -- 0.926 \\
Contradictions & 95\%  & 0.703 -- 0.949 \\
\midrule
Overlaps & 50\%  &  0.902 -- 0.934 \\
Overlaps & 68\%  &  0.866 -- 0.943 \\
Overlaps & 75\%  &  0.852 -- 0.945 \\
Overlaps & 95\%  &  0.804 -- 0.969 \\
\bottomrule
\end{tabularx}
\caption{Statistics for contradictions and overlaps, including counts and similarity score distributions. The 95\% percentile range was chosen for identifying Inter-Case claim contradiction and overlap inconsistencies.}
\label{tab:sim-experiment}
\end{table}

\subsubsection{Negative Claim Generation (Refuted)}
To create the \textsc{Refuted} class, we generated negations of the valid \textsc{Supported} claims using a two-step process designed to produce non-trivial false claims that sound like plausible legal principles. \textsc{Overruled} claims are by nature false and cannot have negations.

\paragraph{Step 1: Plausible Negation}
We first prompted the model (Prompt~\ref{lst:negation_gen}) to generate a plausible but factually incorrect negation of the original claim based on the case facts. This ensured the negation was grounded in the case context rather than being a generic or trivial contradiction, e.g., simply placing \textit{not} in the claim.

\paragraph{Step 2: Principle Generalization}
Initial negations were often lengthy (avg. 212 chars) and contained specific conditional clauses (e.g., ``unless X occurs...''), acting as artifacts that could allow models to distinguish false claims by length or specificity alone. To address this, we applied a refinement LLM pass (Prompt~\ref{lst:negation_len_fix}) instructing the model to rewrite the false claim as a concise, general legal principle. The model was explicitly instructed to remove specific case details and unconditional qualifiers while preserving the (false) core legal assertion. This reduced the average length to 108 chars (see Figure~\ref{fig:negation_lengths}) and ensured the \textsc{Refuted} claims matched the stylistic distribution of the \textsc{Supported} claims.


\begin{table}
\centering
\begin{tabularx}{0.75\linewidth}{lrr}
\toprule
\textbf{Category} & \textbf{Selected (\%)} & \textbf{Total} \\
\midrule
Overall   & 574 (89\%) & 638 \\
Supported & 353 (91\%) & 384 \\
Refuted   & 177 (90\%) & 195 \\
Overruled & 44 (74\%) & 59 \\
\bottomrule
\end{tabularx}
\caption{Quality assessment statistics for the human-annotated test set. Percentages denote the proportion of claims selected as high-quality samples.}
\label{tab:test-set-quality}
\end{table}

\subsection{Human Verification}
The test set consists of 500 claims verified by 2 human annotators. Annotators were instructed to select clear, factual claims free of identifying information. For \textsc{Overruled} claims and claims with multiple ground truth cases, annotators verified the verdict against the full set of cases. From a pool of 638 annotated claims, 574 were validated as high-quality, from which we sampled the final 500. We included an overlap of 100 claims between annotators to calculate the Cohen's Kappa of 0.637, indicating substantial agreement. Table~\ref{tab:test-set-quality} highlights that the \textsc{Overruled} class had a lower human-verification pass rate compared to other classes, indicating the inherent complexity of validating legal overrulings. To further validate the legal quality of the test set, an attorney independently reviewed a sample of 50 claims and found no issues with the claims' factual accuracy or overall quality.

\begin{table}[t]
\centering
\small
\begin{tabular}{lrrrr}
\toprule
\textbf{Dataset} & \textbf{Total} & \textbf{Supported} & \textbf{Refuted} & \textbf{Overruled} \\
\midrule
Train set & 5,794 & 2,605 & 2,732 & 457 \\
Test set & 500 & 280 & 177 & 43 \\
\bottomrule
\end{tabular}
\caption{Overview of the dataset statistics.}
\label{tab:dataset_stats}
\end{table}

\subsection{Dataset Statistics}
The final \textbf{CaseFacts} benchmark consists of a training set of 5,794 claims and a test set of 500 claims. The class distributions are detailed in Table~\ref{tab:dataset_stats}. Each sample shares a unique \texttt{fact\_id} with its corresponding negation and we ensured that no \texttt{fact\_id} overlaps between the training and test sets and between claims and their negations, to prevent data leakage. The claims were randomly sampled when selecting whether to use the \textsc{Supported} claim or its \textsc{Refuted} negation.

In addition to the claims, we release the 3,299 SCOTUS cases used for dataset creation and which serve as ground truth evidence for the claims.



\subsection{Evaluation Metrics}
\label{sec:eval}

We evaluate system performance using a tiered framework designed to assess both retrieval quality and decision-making reliability. All metrics are macro-averaged across claims.

\paragraph{Assessing Retrieval Quality}
To evaluate the quality of retrieved case law, we utilize the \textsc{Evidence Score}. This metric is calculated as the recall score between the predicted and gold case sets. If a system fails to retrieve at least half of the gold supporting cases within the top five results (Recall@5 $< 0.5$), the Evidence Score is penalized to zero. We introduce this threshold because legal research that misses the majority of controlling precedents is functionally useless to a user, regardless of how precise the other retrieved documents might be. This ensures that high scores reflect a robust retrieval of the core legal principles rather than incidental hits.

\paragraph{Assessing Verdict Reliability}
While we report standard \textsc{Verdict Accuracy} (a binary measure of whether the predicted verdict matches the gold label), our primary metric for system ranking is the \textsc{Verdict Score}. This joint, un-weighted metric is computed as the product of the Evidence Score and the Verdict Accuracy.

We prioritize the Verdict Score because it rigorously penalizes ungrounded correctness---instances where a model guesses the correct verdict but cites irrelevant or incorrect cases as evidence. In the high-stakes legal domain, a correct answer derived from faulty reasoning is as dangerous as an incorrect answer. By requiring sufficiently accurate retrieval as a prerequisite for a positive score, this metric ensures that systems are only rewarded when they are right for the right reasons. The k=5 recall also prevents fact-checking systems from using random, incidental case hits as evidence.

\section{Experiments}

\subsection{Naive Factuality Check}

\begin{table}
\centering
\begin{tabularx}{\linewidth}{XXr}
\toprule
\textbf{Evaluation} & \textbf{Judgement} & \textbf{Count} \\
\midrule
\multirow{2}{*}{Standard}
 & Consistent & 8987 (86\%) \\
 & Inconsistent & 1484 (14\%) \\
\midrule
\multirow{3}{*}{Naive}
 & Consistent & 5964 (57\%) \\
 & Inconsistent & 491 (5\%) \\
 & Error & 4016 (38\%) \\
\bottomrule
\end{tabularx}
\caption{Comparison of standard and naive factuality evaluation results, over 10,471 claims, showing counts and proportions of factuality judgment outcomes.}
\label{tab:factuality-results}
\end{table}

When performing the regular factuality check in the process of creating the dataset, we have also ran an experiment where the evidence was not provided to the LLM. We used prompt~\ref{lst:naive_factuality}, and the results are displayed in Table~\ref{tab:factuality-results}. We can observe from this that the LLM is more hesitant to process when evidence is not provided, from the high number of error cases: occurring when the LLM doesn't give the output in the requested format. This is possible due to the LLM's hesitance when instructed to make a decision based on its internal knowledge, when its internal knowledge is not sufficient. In contrast, the LLM never gave outputs in an erroneous format when provided the case's evidence, even though the prompts (Listings \ref{lst:factuality_prompt} and~\ref{lst:naive_factuality}) are quite similar.

\subsection{Similarity Distribution of Cross-Case Claim Inconsistencies}
\label{sec:sim_dist_diff_case}

To locate contradictions and overlapping claims across cases with our compute resource limits, we stratified the space of cross-case claim pairs by their semantic similarity and inspected a uniform subsample from each bin. First, we computed the semantic similarity scores of all possible claim pair combinations, using Qwen3-Embedding-8B~\cite{zhang2025qwen3embeddingadvancingtext}. We used cosine similarity as our similarity measure.

We then performed stratified sampling to estimate the density of inconsistencies across the similarity spectrum. We drew a uniform sample of 1,000 pairs from each 0.1 similarity interval between 0.2 and 0.9. The highest interval (0.9--1.0) contained fewer than 1,000 total pairs; consequently, all pairs in this range were included. Conversely, the 0.0--0.2 interval contained negligible data and was excluded as the pairs were semantically unrelated. For each sampled pair, we employed an LLM-as-a-judge (Prompts~\ref{lst:contra_diff_case_prompt} and~\ref{lst:overlap_diff_case_prompt}) to detect contradictions and overlaps, utilizing the full evidence from both source cases.


\paragraph{Results}
We then inspected the subsample predictions and computed percentile statistics of similarity for contradiction and overlap detections. Both contradictions and overlaps are highly concentrated toward the upper end of the similarity scale: contradictions cluster in the high-similarity region (roughly $\ge$0.7) and overlaps are concentrated even further toward the top (roughly $\ge$0.8), as can be observed in Table~\ref{tab:sim-experiment}. This is inherently logical, as semantically similar claims are more likely to cover similar legal principles.

Overlaps exhibit a substantially higher positive-rate within their window than contradictions, while contradictions are rarer. This necessitated the additional resolution step with the filter by disposition for the contradictions. These results justify restricting the LLM-judge to a high-similarity slice of the pairwise space: doing so covers the large majority of contradictions and overlaps while keeping compute cost manageable.

\subsection{Baseline Experiments}

We evaluate two LLM baselines for Supreme Court claim verification that differ only in their access to external information at inference time. Both baselines use the same prompt (\ref{lst:baseline_prompt}), evaluation, and LLM. GPT-4o was selected as the LLM to avoid self-preference bias, as it is not part of the Qwen3 or Gemma3 model families~\cite{wataoka2025selfpreferencebiasllmasajudge}.

The prompt presents the model with a legal claim and instructs it to determine the verdict against U.S. Supreme Court case law, with the provided verdict definitions. In addition to predicting a verdict, the model is required to identify the Supreme Court cases that justify the decision and to rank them by importance. To ensure a balanced evaluation, the prompt includes an explicit list of valid Supreme Court case names and instructs the model to cite only from this list. This consists of the cases that were used to create the dataset. 

\paragraph{Baseline Setting}
There are two settings for the baseline system: Naive (no-search), and Search enabled. Both baselines were evaluated on the 3 metrics described in Section~\ref{sec:eval}. In the naive setting, the model receives only the prompt. The model must rely solely on its internal knowledge to determine the correct verdict and select the cases to be used as evidence. This baseline evaluates the model’s ability to perform reasoning and case recall without external evidence retrieval. In the search setting, the model is additionally allowed to perform web search during inference, using OpenAI's web search tool\footnote{\url{https://platform.openai.com/docs/guides/tools-web-search}}. This enables the model to consult external sources when performing the fact-checking task.

\paragraph{Results}

From Table~\ref{tab:llm-search-comparison}, it is evident that the major challenge for this benchmark dataset is gathering evidence, as the evidence score (case recall metric) is much lower than the verdict accuracy. This points to the verdicts being easier to predict by the LLM, as both search baselines perform similarly on the verdict prediction. However, for a fact-checking application, the quality of the evidence retrieved is quite important for users' trustworthiness~\cite{anand2022explainableinformationretrievalsurvey, Schlichtkrull2023AVeriTeCAD}, hence why we use the composite metric of ``verdict score'' as our primary metric for this dataset.

The Naive baseline outperforms the Search baseline on all three metrics, as can be observed from Table~\ref{tab:llm-search-comparison}. This pattern indicates that allowing unrestricted web search did not improve, and in fact degraded, both evidence retrieval quality and the joint evidence‑weighted verdict performance. A primary contributor is that web search sometimes surfaced cases outside the 3,299 case list or returned noisy, tangentially relevant cases, which led to lower overlap with the gold supporting cases therefore penalties by the evaluation metrics. By contrast, the Naive setting, relying on the model’s internal knowledge and the constrained case list, produced more consistent case selections and higher overall scores, although not by much. These results suggest that any retrieval augmentation should be tightly constrained to the validated case set and paired with stronger re-ranking or filtering.


\begin{table}
\centering
\begin{tabularx}{\linewidth}{Xcc}
\toprule
\textbf{Model} & \textbf{Metric} & \textbf{Performance} \\
\midrule
Naive & Evidence Score & \textbf{0.232} \\
Naive & Verdict Accuracy & \textbf{0.600} \\
Naive & Verdict Score & \textbf{0.185} \\
\midrule
Search & Evidence Score & 0.212 \\
Search & Verdict Accuracy & 0.574 \\
Search & Verdict Score & 0.169 \\
\bottomrule
\end{tabularx}
\caption{Comparison of evidence and verdict metrics for LLM configurations with and without search.}
\label{tab:llm-search-comparison}
\end{table}


\subsection{Finetuning Semantic Similarity Models}

\begin{table*}
\centering
\small
\begin{tabularx}{\textwidth}{@{} l c *{3}{>{\centering\arraybackslash}X} @{}}
\toprule
\textbf{Model} & \textbf{Epochs} & \textbf{Recall@1} & \textbf{Recall@5} & \textbf{Recall@10} \\
\midrule
BM25    & --  & 0.1160 & 0.2180 & 0.2500 \\
ColBERT & --  & 0.2560 & 0.4620 & 0.5420 \\
\midrule
\multirow{3}{*}{bge-base-en-v1.5}
  & 1 & 0.2195 \textrightarrow\ 0.3143 \gain{+.0948} & 0.3957 \textrightarrow\ 0.5941 \gain{+.1984} & 0.4623 \textrightarrow\ 0.6788 \gain{+.2166} \\
  & 3 & 0.2195 \textrightarrow\ 0.3373 \gain{+.1178} & 0.3957 \textrightarrow\ 0.6100 \gain{+.2143} & 0.4623 \textrightarrow\ 0.6748 \gain{+.2126} \\
  & 5 & 0.2195 \textrightarrow\ 0.3323 \gain{+.1128} & 0.3957 \textrightarrow\ 0.6009 \gain{+.2052} & 0.4623 \textrightarrow\ 0.6758 \gain{+.2136} \\
\midrule
\multirow{3}{*}{MiniLM-L6-v2}
  & 1 & 0.1452 \textrightarrow\ 0.2322 \gain{+.0870} & 0.3231 \textrightarrow\ 0.4809 \gain{+.1578} & 0.3969 \textrightarrow\ 0.5626 \gain{+.1656} \\
  & 3 & 0.1452 \textrightarrow\ 0.2358 \gain{+.0906} & 0.3231 \textrightarrow\ 0.5212 \gain{+.1981} & 0.3969 \textrightarrow\ 0.6269 \gain{+.2300} \\
  & 5 & 0.1452 \textrightarrow\ 0.2648 \gain{+.1196} & 0.3231 \textrightarrow\ 0.5305 \gain{+.2075} & 0.3969 \textrightarrow\ 0.6529 \gain{+.2560} \\
\midrule
\multirow{3}{*}{Qwen3-Embed-0.6B}
  & 1 & 0.2952 \textrightarrow\ 0.4999 \gain{+.2047} & \textbf{0.5415 \textrightarrow\ 0.7930} \gain{\textbf{+.2515}} & 0.6471 \textrightarrow\ 0.8382 \gain{+.1911} \\
  & 3 & \textbf{0.2952 \textrightarrow\ 0.5364} \gain{\textbf{+.2412}} & 0.5415 \textrightarrow\ 0.7652 \gain{+.2237} & 0.6471 \textrightarrow\ 0.8368 \gain{+.1897} \\
  & 5 & 0.2952 \textrightarrow\ 0.5240 \gain{+.2288} & 0.5415 \textrightarrow\ 0.7669 \gain{+.2254} & \textbf{0.6471 \textrightarrow\ 0.8524} \gain{\textbf{+.2053}} \\
\bottomrule
\end{tabularx}
\caption{Recall@$k$ comparisons (base \textrightarrow\ fine-tuned) for baseline and dense retrieval models across training epochs. Parenthesized values show improvement over the base model. The highest accuracy per metric is \textbf{bolded}.}
\label{tab:recall-comparisons}
\end{table*}

In addition to prompting-based LLM baselines, we fine-tune dense embedding models for mapping legal claims to the relevant SCOTUS cases, with CaseFacts' training set. This experiment isolates the retrieval component of the broader fact-checking task and assesses whether supervised representation learning improves the ability to identify legally relevant precedents.

Given a legal claim from CaseFacts as a query, the model must retrieve the Supreme Court cases that are the ground truth evidence for that claim. Each case is represented by its textual description, consisting of the case facts, questions and conclusions. Performance is measured using the test set, and the metrics are $\mathrm{Recall}@k$ for $k \in \{1, 5, 10\}$.

We evaluated three base encoder families (BGE~\cite{bge_embedding}, Qwen3‑Embedding‑0.6B~\cite{zhang2025qwen3embeddingadvancingtext}, and MiniLM~\cite{wang2020minilmdeepselfattentiondistillation}) and compared each base model to versions fine‑tuned with MultipleNegativesRankingLoss for 1, 3, and 5 epochs. Fine-tuning encourages the model to embed claims closer to their corresponding cases while pushing apart unrelated cases, aligning the embedding space with the structure of the dataset.

\paragraph{Results}
Table \ref{tab:recall-comparisons} reports that, as expected, across all models and epochs, supervised fine-tuning yields substantial improvements over the corresponding pretrained baselines. Fine-tuning consistently improves Recall@1, indicating that training substantially increases the likelihood that the most relevant Supreme Court case is ranked first. This is particularly pronounced for stronger and larger base models: Qwen3-Embedding-0.6B shows the largest single-step improvement at Recall@1 (+24.1 points after three epochs), while bge-base-en-v1.5 and MiniLM-L6-v2 exhibit smaller, but steady gains as training epochs increase. This pattern of increasing gains for more epochs holds for the smaller models as well.

This demonstrates the usefulness of CaseFacts' training set for supervised fine-tuning of the retrieval portion, displaying that trained retrieval models can be complementary to an LLM-based fact-checking system. This approach can be an alternative to naive and web-search LLM fact-checkers in the legal domain.


\subsection{LLM-Judge Reliability}

To evaluate the reliability of the LLM-as-a-judge steps, we have repeated the factuality, cross-case contradiction and overlap tasks on the human-annotated claims from the test set. As the issues the inconsistencies identified by the aforementioned tasks should not be present in the test set, this experiment evaluates alignment with human annotations. The results in Table~\ref{tab:llm-judge-eval} indicate that the LLM is very consistent with human judgment on the sub-tasks.

\begin{table}
\centering
\begin{tabularx}{\linewidth}{Xrrr}
\toprule
\textbf{Metric} & \textbf{Accuracy} & \textbf{Correct} & \textbf{Incorrect} \\
\midrule
Factuality     & 97.4\% & 487     & 13     \\
Contradiction  & 95.5\% & 238,148 & 11,352 \\
Overlap        & 97.4\% & 243,113 & 6,387  \\
\midrule
Total          & 96.5\% & 481,748 & 17,752 \\
\bottomrule
\end{tabularx}
\caption{Evaluation of LLM-judge across various tasks.}
\label{tab:llm-judge-eval}
\end{table}

\subsection{Effectiveness of Private Information Removal}
To ascertain the effectiveness of the private information removal in Section~\ref{sec:claim-gen}, we evaluated our dataset's claims with \texttt{roberta-large-ner-english} and found that approximately 25\% of the claims mentioned an entity. Upon further observation, we found that most of these were ``The United States'' or some legal entity/statute, which we felt were capturing undesirable entities. We used an LLM judge to check the claims and found that only 154 (2.5\%) of the claims contained a relevant entity.

To assess whether these claims posed a problem for evaluation, we compared baseline performance on this entity-containing subset against the test set. The results for both Search-enabled and Naive show no meaningful difference; for example, under the Naive setting, \textsc{Evidence Score} was 0.232 on the entity subset versus 0.222 on the test, and \textsc{Verdict Score} was 0.222 versus 0.185. We therefore retain these claims in the dataset, as they may better reflect a real-world distribution where a minority of legal claims do reference specific identifying information, and this subset remains small enough (2.5\%) to have negligible impact on overall benchmark statistics.


\section{Related Work}

\paragraph{Legal Reasoning and Benchmarks.}
Recent efforts have established robust benchmarks to evaluate Large Language Models (LLMs) on legal tasks. LegalBench~\cite{guha2023legalbenchcollaborativelybuiltbenchmark} provides a comprehensive suite of tasks ranging from clause classification to rule application. While LegalBench covers various aspects of legal reasoning, none of its constituent tasks directly tackle the problem of automated fact-checking, where a system must verify an external claim against a corpus. Our work can be viewed as a synthesis of multiple LegalBench reasoning skills---requiring both the retrieval of relevant precedent and the entailment capabilities necessary to verify a claim.

Similarly, CaseHOLD~\cite{zheng2021doespretraininghelpassessing} focuses on identifying case holdings---the governing legal rules applied to specific facts that serve as binding precedent. However, this benchmark operates strictly within the formal legal register, using judicial citations as prompts rather than colloquial claims. Additionally, it frames the task as multiple-choice matching where distractors are simply irrelevant precedents selected via TF-IDF, rather than the adversarial distortions found in real-world misinformation. The Competition on Legal Information Extraction/Entailment~\cite{COLIEE} similarly addresses case law retrieval and entailment, but operates over formal legal texts in a formal-to-formal setting. Our benchmark addresses these gaps by verifying informal assertions against formal rulings, testing resilience against subtle misinterpretations rather than just formal document retrieval precision.

\paragraph{The Semantic Gap in Legal Retrieval}
A primary challenge in legal fact-checking is the linguistic disparity between colloquial claims and formal legal texts. \citet{chen2025-retrieving} and \citet{mori2025assessingperformancegaplexical} investigate this friction, highlighting the difficulty of retrieving relevant legal provisions using informal or layperson queries. This issue is further compounded by the structural complexity of legal documents; \citet{reuter-etal-2025-towards} identify that lexical redundancy and fragmented information in legal datasets cause standard RAG systems to fail frequently. Similarly, \citet{sym17050633} argue that traditional semantic similarity metrics often miss the nuanced relevance of statutes, requiring dynamic adaptation rather than static retrieval. The limited performance observed in these studies underscores the difficulty of the task and motivates our focus on bridging the gap between informal political discourse and the technical language of SCOTUS opinions.

\paragraph{Domain-Specific Fact-Checking}
There is a growing paradigm shift in automated fact-checking, moving away from ``one-size-fits-all'' systems~\cite{putta2025claimcheckrealtimefactcheckingsmall,Braun2024DEFAMEDE} toward domain-specific architectures. \citet{wang-etal-2025-openfactcheck} and \citet{devasier-etal-2025-task} argue that specialized domains---such as congressional law or medicine---require distinct pipelines that general-purpose systems cannot support. \citet{hu-etal-2025-fine} recently addressed this in the context of Legal Question Answering by fine-tuning models to reduce hallucinations; however, their work focuses on generating truthful answers to questions, whereas our work focuses on the task of verifying external claims against precedents.

To address the scarcity of real-world training data in these specialized domains, recent works have successfully employed synthetic data generation. \citet{zhao-flanigan-2025-synthverify} and \citet{zhang2025enhancinghealthfactcheckingllmgenerated} demonstrate that creating synthetic factual claims from source texts can significantly improve performance on downstream verification tasks. This methodology is supported by broader surveys in the field, such as \citet{Nad__2025}, which validate the use of LLMs to generate high-quality training data in domains constrained by data scarcity and privacy. Following this methodology, we leverage LLMs to generate diverse claim-evidence pairs.
\section{Conclusion}

In this work, we introduced \textbf{CaseFacts}, a benchmark designed to address the unique challenges of automated fact-checking in the legal domain. By bridging the semantic gap between colloquial assertions and formal Supreme Court precedents, CaseFacts moves beyond traditional ``legalese-to-legalese'' tasks, offering a more realistic evaluation of how legal information is consumed in public discourse and social media. A key contribution of our work is the rigorous modeling of precedent validity through the \textsc{Overruled} class. By developing a scalable pipeline to identify and verify overrulings, we demonstrate that legal truth is not static and that robust verification systems must account for the dynamic evolution of case law and the formation of new legal precedent.

Our experiments reveal significant limitations in current state-of-the-art models. We observed that providing LLMs with unrestricted web search often degrades performance due to the retrieval of noisy or irrelevant case law, highlighting the need for retrieval systems that prioritize authoritative and precise evidence. Furthermore, while fine-tuning embedding models on our synthetic training data yielded substantial gains in retrieval, the gap between simple verdict prediction and evidence-grounded verification remains large. This opens the door to creation of specialized systems that can better navigate the hierarchical and temporal structures of legal evidence.


\section*{Limitations}
First, CaseFacts relies on expert-synthesized summaries from Oyez rather than full judicial opinions. While this ensures high-quality structured evidence, it abstracts away the complex, long-context reasoning found in raw legal texts, potentially simplifying retrieval compared to a full-text corpus.

Our heuristic approach to identifying overruled cases relies on semantic similarity thresholds. While necessary for computational feasibility, this method inherently misses some cases with low semantic similarity, though we expect the number to be quite small.

While we employ a rigorous pipeline to generate hard negative refuted claims, they remain synthetically derived. They may not fully capture the nuance of human-generated legal misinformation, which often relies on subtle misinterpretation or out-of-context citation rather than the direct negation of legal principles.

Finally, CaseFacts focuses exclusively on U.S. Supreme Court rulings. It does not account for statutory law, regulatory codes, or state-level jurisprudence, which are frequent targets of misinformation in the broader legal domain. This also means that if supreme court rulings are overridden by congressional legislation, we are unable to label them as overruled. Furthermore, this work does not consider legal precedence in other countries.

\section*{Ethical Concerns}

\paragraph{No Legal Advice and Risk of Harm}
While CaseFacts aims to improve the reliability of automated fact-checking in the legal domain, the models trained or evaluated on this benchmark are research prototypes and should not be used as a substitute for professional legal counsel. Legal AI systems are prone to hallucinations, and in high-stakes environments, such errors can lead to severe consequences, including financial loss or the deprivation of rights. We emphasize that any downstream application derived from this work should utilize a human-in-the-loop workflow, particularly when verifying claims that influence real-world legal decision-making.

\paragraph{Bias and Harmful Historical Content}
Our dataset is derived from U.S. Supreme Court rulings via Oyez summaries. Historically, judicial corpora reflect the prejudices of their time. Consequently, the dataset contains assertions regarding slavery, segregation, and disenfranchisement (e.g., claims stating that enslaved persons are property). While these claims are labeled as \textsc{Overruled} or contextually specific to valid historical precedents, they represent morally wrong viewpoints which we do not agree with. We retain them because a robust legal fact-checker must be able to identify that such precedents---though once "the law"---are no longer valid. Users should be aware that the dataset contains text describing human rights violations.

\paragraph{Dual Use and Misinformation}
Our methodology includes a specific pipeline for generating hard negatives. While this is necessary to train robust discriminators, the same techniques could theoretically be misused to generate convincing legal disinformation at scale. However, we believe the benefit of releasing a dataset to detect such misinformation outweighs the risk.

\paragraph{Privacy and Data Usage}
Our claim generation pipeline explicitly prompts models to remove specific identifiers, such as party names and dates, to focus on general legal principles. This minimizes the risk of generating claims that unintentionally expose sensitive information regarding private individuals involved in historical litigation. All source data is derived from publicly available summaries provided by Oyez.

\bibliography{custom}

\appendix

\section{Appendix}

\subsection{Reproducibility}
All of our code is open-sourced at \url{https://github.com/idirlab/supreme-court-dataset}. Github Copilot was used in the creation of some of our code.

\subsection{Token Length Distribution}

\begin{figure*}
  \centering
  \includegraphics[width=0.75\linewidth]{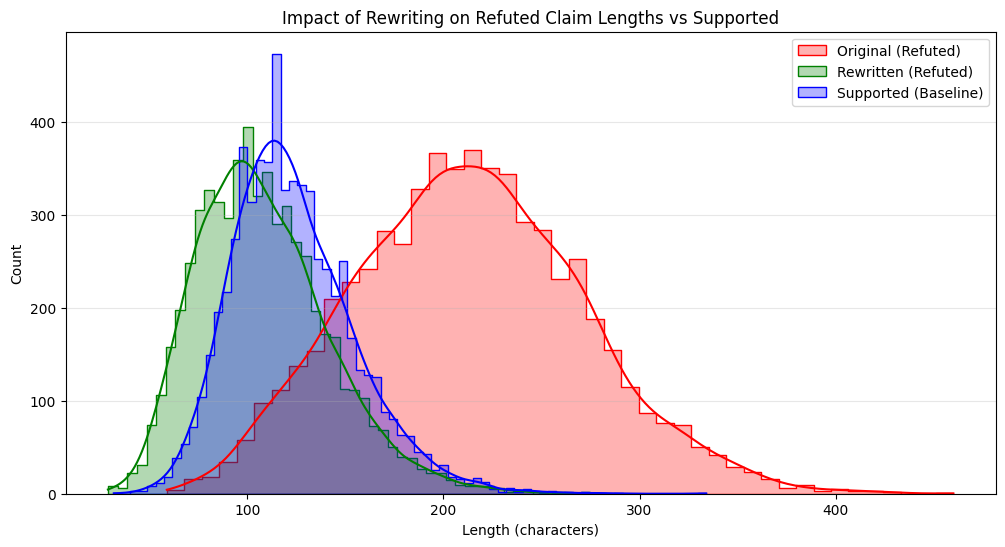}
  \caption{\label{fig:negation_lengths}Graph displaying the changes in character length before and after the LLM pass with prompt~\ref{lst:negation_len_fix}. The length of the supported claims, which negations are generated from, is provided for comparison.}
\end{figure*}

Figure \ref{fig:negation_lengths} provides an analysis of the token length distributions for the claims in our dataset. It compares the lengths of the originally generated valid claims against the generated negations to demonstrate that the negation length-fixing step (Listing \ref{lst:negation_len_fix}) successfully maintained comparable claim complexity.

\subsection{LLM Prompts}
\label{app:prompts}

This section details the specific prompts used to construct the dataset and run baseline experiments.

\paragraph{Claim Generation and Factuality}
Listing \ref{lst:claim_gen_prompt} displays the prompt used to extract atomic legal claims from case summaries. To ensure these claims are supported by the text, we utilize a factuality checker (Listing \ref{lst:factuality_prompt}). We also include a ``naive'' factuality prompt (Listing \ref{lst:naive_factuality}) used for ablation studies where the model judges factuality without access to the case evidence.

\paragraph{Intra-Case Relationships}
For claims derived from the same case, we verify their relationship to ensure diversity and consistency. We detect contradictions using Listing \ref{lst:contra_same_case_prompt} and semantic overlaps (redundancy) using Listing \ref{lst:overlap_same_case_prompt}. If conflicts or redundancies are found, they are resolved using the prompts in Listing \ref{lst:contra_same_case_resolution_prompt} and Listing \ref{lst:overlap_same_case_resolution_prompt}, respectively.

\paragraph{Inter-Case Relationships}
We perform similar consistency checks for claims originating from different cases. Listings \ref{lst:contra_diff_case_prompt} and \ref{lst:contra_diff_case_res_prompt} show the prompts for detecting and resolving contradictions (e.g., overruling) between cases. Listings \ref{lst:overlap_diff_case_prompt} and \ref{lst:overlap_diff_case_res_prompt} detail the detection and resolution of cross-case redundancies.

\paragraph{Negation Generation}
To create ``Refuted'' samples, we generate plausible negations of valid claims using the prompt in Listing \ref{lst:negation_gen}. We then apply a length-fixing and style-refinement prompt (Listing \ref{lst:negation_len_fix}) to ensure the negated claims remain concise and comparable to the original claims.

\paragraph{Baselines}
Finally, Listing \ref{lst:baseline_prompt} provides the prompt used for the zero-shot baseline predictions, where the model must retrieve evidence and output a verdict based on a provided list of valid cases.


\lstset{
    basicstyle=\ttfamily\small,
    backgroundcolor=\color{gray!10}, 
    linewidth=\linewidth,
    breaklines=true,
    frame=single, 
    rulecolor=\color{black}, 
    showstringspaces=true, 
    numbers=left, 
    numberstyle=\tiny\color{gray}, 
    xleftmargin=0em, 
    numbers=none, 
    columns=fullflexible,
    breakindent=0pt,
    literate={`}{\textasciigrave}1,
    moredelim=**[is][\color{gray!70}]{@}{@} 
}
\begin{figure*}[th]

\begin{lstlisting}[caption={Prompt for Claim Generation}, label=lst:claim_gen_prompt]
# Instructions:
Generate truthful, factual claims about the legal implications of this case using simple, everyday language.
Avoid legal jargon or court-report phrasing. Do not use phrases like "the court found," "the decision clarified," or "under this test."
Write claims as clear statements of what the law allows or does not allow.
Adhere to the following rules:
- Do not use any part of the case name or other identifying information in the claim.
- Do not make claims that only apply to the parties in this case; claims must be general legal principles.
- Keep each claim focused on one central idea and make it easy to understand.
- Each claim must be distinct from the others; do not repeat the same core idea.
- Use direct, plain wording rather than legal formulations.

Output Format:
## Return the claim in a JSON object with the following format:
```json
{{
    "claim1": "...",
    "claim2": "...",
    ...
}}
```

## Facts:
{facts}
 
# Question:
{question}

# Conclusion:
{conclusion}

\end{lstlisting}
\end{figure*}

\begin{figure*}[th]

\begin{lstlisting}[caption={Prompt for the Factuality Check}, label=lst:factuality_prompt]
You are a legal expert. Read a short claim about a supreme court case plus the case facts, question, and conclusion. Decide whether the claim is factually consistent with the evidence from the case or is factually inconsistent with the case evidence.

Rules:
1. Base your judgment only on the Supreme Court evidence.
2. If the evidence does not support the claim, do not label as consistent.
3. Do not rely on outside knowledge or assumptions.
4. Do not invent information that is not in the evidence.

Claim: {claim}

Case Evidence:
Facts: {facts}
Question: {question}
Conclusion: {conclusion}

## Output Format:
Return a JSON object in the following format:
```json
{{
    "explanation": "...",
    "contradiction": "<consistent/inconsistent>",
    ...
}}
```

\end{lstlisting}
\end{figure*}

\begin{figure*}[th]

\begin{lstlisting}[caption={Prompt for the Contradiction Check within the same case}, label=lst:contra_same_case_prompt]
You are a legal expert. Read two short claims about a case plus the case facts, question, and conclusion. Decide whether the two claims are contradicting (negation or opposite entailment), or consistent (when they are unrelated or entail).
You must output an explanation for your decision in the "explanation" field. Then, also provide a decision in the "contradiction" field: "contradiction" if the claims are contradicting, and "consistent" if they are not.

Claim 1: {claim1}
Claim 2: {claim2}

Case Evidence:
Facts: {facts}
Question: {question}
Conclusion: {conclusion}

## Output Format:
Return a JSON object in the following format:
```json
{{
    "explanation": "...",
    "contradiction": "<contradiction/consistent>",
    ...
}}
```

\end{lstlisting}
\end{figure*}

\begin{figure*}[th]

\begin{lstlisting}[caption={Prompt for the Overlap Check within the same case}, label=lst:overlap_same_case_prompt]
You are a legal expert. Read two short claims about a case plus the case facts, question, and conclusion. Decide whether the two claims are saying the same thing (being redundant) or are meaningfully different regarding their meaning.
You must output an explanation for your decision in the "explanation" field. Then, also provide a decision in the "overlap" field: "redundant" if the claims are redundant, and "different" if they are not.

Claim 1: {claim1}
Claim 2: {claim2}

Case Evidence:
Facts: {facts}
Question: {question}
Conclusion: {conclusion}

## Output Format:
Return a JSON object in the following format:
```json
{{
    "explanation": "...",
    "overlap": "<redundant/different>",
    ...
}}

\end{lstlisting}
\end{figure*}

\begin{figure*}[th]

\begin{lstlisting}[caption={Prompt for Contradiction resolutions within the same case}, label=lst:contra_same_case_resolution_prompt]
You are a legal expert. You are given two claims about the same Supreme Court case that have been identified as contradictory. Read the two short claims about the case and the case facts, question, and conclusion.
Your task is to determine which claim is factually correct based on the evidence, or if neither is correct. In the "decision" field, only return one of the decision categories.

Claim 1: {claim1}
Claim 2: {claim2}

Reason for contradiction: {explanation}

Case Evidence:
Facts: {facts}
Question: {question}
Conclusion: {conclusion}

Analyze the evidence and decide:
1. Is Claim 1 correct and Claim 2 incorrect?
2. Is Claim 2 correct and Claim 1 incorrect?
3. Are both claims incorrect?
4. Are both claims partially correct but phrased poorly? (If so, provide a merged/corrected claim).

## Output Format:
Return a JSON object in the following format:
```json
{{
    "explanation": "...",
    "decision": "<claim1_correct | claim2_correct | neither_correct | both_partial>",
    "corrected_claim": "..." (optional, if decision is 'both_partial')
}}
```

\end{lstlisting}
\end{figure*}

\begin{figure*}[th]

\begin{lstlisting}[caption={Prompt for Overlap resolutions within the same case}, label=lst:overlap_same_case_resolution_prompt]
You are a legal expert. You are given two claims about the same Supreme Court case that are redundant (saying the same thing). Read the two short claims about the case and the case facts, question, and conclusion. 
Your task is to choose the one that is better written, more precise, or more comprehensive. You are also given the reason for redundancy, take that into account when making your decision. In the "keep" json field, return the strings "claim1" or "claim2", not the actual content of the claims.

Claim 1: {claim1}
Claim 2: {claim2}

Reason for redundancy: {explanation}

Case Evidence:
Facts: {facts}
Question: {question}
Conclusion: {conclusion}

## Output Format:
Return a JSON object in the following format:
```json
{{
    "explanation": "...",
    "keep": "<claim1/claim2>"
}}
```

\end{lstlisting}
\end{figure*}

\begin{figure*}[th]

\begin{lstlisting}[caption={Prompt for the Contradiction Check within different cases}, label=lst:contra_diff_case_prompt]
You are a legal expert. Read two short claims from different cases and the two cases' associated facts, legal questions, and conclusions.
Decide whether the two claims are contradicting (negation or opposite entailment), or consistent (when they are unrelated or entail).
You must output an explanation for your decision in the "explanation" field. Then, also provide a decision in the "contradiction" field: "contradiction" if the claims are contradicting, and "consistent" if they are not.

## Output Format:
Return a JSON object in the following format:
```json
{{
    "explanation": "...",
    "contradiction": "<contradiction/consistent>",
    ...
}}
```

Claim 1: {claim1}
Claim 2: {claim2}

Claim 1 Case Evidence:
Facts: {facts1}
Legal Question: {api_question1}
Conclusion: {api_conclusion1}

Claim 2 Case Evidence:
Facts: {facts2}
Legal Question: {api_question2}
Conclusion: {api_conclusion2}

\end{lstlisting}
\end{figure*}

\begin{figure*}[th]

\begin{lstlisting}[caption={Prompt for the Overlap Check within different cases}, label=lst:overlap_diff_case_prompt]
You are a legal expert. Read two short claims from different cases and the two cases' associated facts, legal questions, and conclusions.
Decide whether the two claims are saying the same thing (being redundant) or are meaningfully different regarding their meaning.
You must output an explanation for your decision in the "explanation" field. Then, also provide a decision in the "overlap" field: "redundant" if the claims are redundant, and "different" if they are not.

## Output Format:
Return a JSON object in the following format:
```json
{{
    "explanation": "...",
    "overlap": "<redundant/different>",
    ...
}}
```

Claim 1: {claim1}
Claim 2: {claim2}

Claim 1 Case Evidence:
Facts: {facts1}
Legal Question: {api_question1}
Conclusion: {api_conclusion1}

Claim 2 Case Evidence:
Facts: {facts2}
Legal Question: {api_question2}
Conclusion: {api_conclusion2}

\end{lstlisting}
\end{figure*}

\begin{figure*}[th]

\begin{lstlisting}[caption={Prompt for the Contradiction Resolution within different cases}, label=lst:contra_diff_case_res_prompt]
You are a legal expert. You are given two claims from different Supreme Court cases that have been identified as contradictory. Read the two short claims about the cases and the two cases' facts, question, and conclusion.
1. Are they overruling one another? Indicate "case1_overruled" if Case 2's evidence points to overruling Case 1's evidence. Indicate "case2_overruled" if Case 1's evidence points to overruling Case 2's evidence. Take into account their ruling dates when making overruling decisions.
2. Are they consistent given context? (e.g. different jurisdictions, different specific facts). Indicate "consistent" in the decision field. Even if the claims are slightly contradicting, they are consistent as long as they propagate different legal principles. This will be quite common, as true overruling contradictions are rare in Supreme Court cases.

Claim 1: {claim1}
Claim 2: {claim2}

Case 1 Evidence:
Ruling Date: {date1}
Facts: {facts1}
Question: {api_question1}
Conclusion: {api_conclusion1}


Case 2 Evidence:
Ruling Date: {date2}
Facts: {facts2}
Question: {api_question2}
Conclusion: {api_conclusion2}

Output JSON:
{{
    "explanation": "...",
    "decision": "case1_overruled" | "case2_overruled" | "consistent",
}}
```

\end{lstlisting}
\end{figure*}

\begin{figure*}[th]

\begin{lstlisting}[caption={Prompt for the Overlap Resolution within different cases}, label=lst:overlap_diff_case_res_prompt]
You are a legal expert. You are given two claims from different cases that have been identified as redundant (overlapping).
Your task is to decide how to resolve this overlap. You can:
1. Keep Claim 1 (if it is more accurate, comprehensive, or better phrased). The ground truth for Claim 1 will be both cases.
2. Keep Claim 2 (if it is more accurate, comprehensive, or better phrased). The ground truth for Claim 2 will be both cases.
3. Merge them (create a new claim that combines the information from both).

Claim 1: {claim1}
Claim 2: {claim2}

Case 1 Evidence:
Facts: {facts1}
Question: {api_question1}
Conclusion: {api_conclusion1}

Case 2 Evidence:
Facts: {facts2}
Question: {api_question2}
Conclusion: {api_conclusion2}

Output JSON:
{{
    "reasoning": "...",
    "decision": "keep_1" | "keep_2" | "merge",
    "merged_claim": "..." (only if decision is merge, otherwise null)
}}

\end{lstlisting}
\end{figure*}

\begin{figure*}[th]

\begin{lstlisting}[caption={Prompt for the Negations Generation}, label=lst:negation_gen]
You are a legal expert. Read the following claim about a legal case, along with the case's facts, question, and conclusion.
Your task is to generate a negation of this claim. The negation should be plausible but factually incorrect based on the original claim and the case evidence.
Provide an explanation for why this negation contradicts the original claim in the "explanation" field. Then, provide the negated claim in the "negation" field.

Claim: {claim}

Case Evidence:
Facts: {facts}
Question: {question}
Conclusion: {conclusion}

## Output Format:
Return a JSON object in the following format:
```json
{{
    "explanation": "...",
    "negation": "..."
}}
```
\end{lstlisting}
\end{figure*}

\begin{figure*}[th]

\begin{lstlisting}[caption={Prompt for the Negations Length Fixing}, label=lst:negation_len_fix]
# Instructions:
Rewrite the following legal claim to be a concise, general legal principle.
The input claim is a "refuted" (false) legal claim, but it is currently might be too long, specific, or conditional. The case the claim originated from is provided as context. It will contradict the claim, do not change the meaning of the claim.
Your task is to rewrite it so it is:
1. Independent of specific case details or parties (remove names, dates, specific locations).
2. Unconditional (remove "unless", "especially if", or specific factual caveats).
3. Concise and direct (simple, everyday language).
4. Focused on the core legal principle being asserted (even if that principle is false).

You must not change the meaning of the claim. If some details are necessary to preserve the meaning, keep them, even if that makes the claim lengthy.

If the claim is already concise and general, you may return it as is or with minor improvements.

## Input Claim:
"{claim}"

## Output Format:
Return a JSON object with a single key "rewritten_claim":
```json
{{
    "rewritten_claim": "..."
}}
\end{lstlisting}
\end{figure*}

\begin{figure*}[th]

\begin{lstlisting}[caption={Prompt for the Naive Factuality check}, label=lst:naive_factuality]
You are a legal expert. Read a short claim about a supreme court case. Decide whether the claim is factually consistent with the details of the case.

Rules:
1. Base your judgment on your internal knowledge of the Supreme Court case.
2. If you are unsure or do not know the case, do not label as consistent.

Claim: {claim}

## Output Format:
Return a JSON object in the following format:
```json
{{
    "explanation": "...",
    "contradiction": "<consistent/inconsistent>",
    ...
}}
```
\end{lstlisting}
\end{figure*}

\begin{figure*}[th]

\begin{lstlisting}[caption={Prompt for Baseline Predictions}, label=lst:baseline_prompt]
You are a legal expert. Your task is to analyze a legal claim and determine its veracity based on US Supreme Court cases.
You must determine if the claim is "Supported", "Refuted", or "Overruled" by the case law.
You must also identify the specific Supreme Court cases that serve as evidence for your decision. List them in order of importance (most important first). You will be penalized for irrelevant and incorrect citations, so prioritize accuracy and conciseness of citations.

Constraints:
1. You must ONLY cite cases from the provided list of valid Supreme Court cases. Do not invent cases or cite cases not in the list.
2. Do not guess. If you are unsure, provide your best estimate but prioritize accuracy.
3. Output must be a valid JSON object.

Valid Supreme Court Cases:
{case_list}

Claim: {claim}

Respond with a JSON object in the following format:
{{
    "explanation": "Brief explanation of your reasoning.",
    "cases": ["Case Name 1", "Case Name 2", ...],
    "verdict": "Supported" or "Refuted" or "Overruled"
}}
\end{lstlisting}
\end{figure*}


\end{document}